# Assisting Blind People Using Object Detection with Vocal Feedback

Heba Najm
Faculty of Information Technology
University of Benghazi
Benghazi, Libya
heba.hnajem@gmail.com

Khirallah Elferjani
Faculty of Information Technology
University of Benghazi
Benghazi, Libya
khirallah.elfarjani@uob.edu.ly

Alhaam Alariyibi
Faculty of Information Technology
University of Benghazi
Benghazi, Libya
Alhaam.Alariyibi@uob.edu.ly

*Abstract*— For visually impaired people, it is highly difficult to make independent movement and safely move in both indoors and outdoors environment. Furthermore, these physically and visually challenges prevent them from in day-to-day live activities. Similarly, they have problem perceiving objects of surrounding environment that may pose a risk to them. The proposed approach suggests detection of objects in real-time video by using a web camera, for the object identification, process. You Look Only Once (YOLO) model is utilized which is CNN-based real-time object detection technique. Additionally, The OpenCV libraries of Python is used to implement the software program as well as deep learning process is performed. Image recognition results are transferred to the visually impaired users in audible form by means of Google text-to-speech library and determine object location relative to its position in the screen. The obtaining result was evaluated by using the mean Average Precision (mAP), and it was found that the proposed approach achieves excellent results when it compared to previous approaches.

## I. INTRODUCTION

One of the most important human senses is vision and it makes human's life more easier in surrounding environment. According to a statistical analysis study of the World Health Organization (WHO)[1], there are about 285 million people around the world who suffer from a vision impairment or total blindness. These visually impaired people usually face many challenges in recognizing the things around them and avoiding the difficulties they face in performing their daily activities. Traditionally, the usage of guide canes is considered the solution to these situations. However, many visually impaired people are preferred due to its user-friendly and cost-efficient. Unfortunately, a guide cane is not enough for blind or visually impaired people to be independent and come over their difficulties in their daily life. They still need someone to help them find an object, cross the street, navigate routes, and give directions. Therefore, visually impaired people cannot predict the exact environment features or know exactly what types of objects lies in front them using only conventional methods.

There are varieties of popular computer vision techniques e.g. image classification, semantic segmentation, object tracking, and object detections. Computer vision uses deep learning algorithms to perform its tasks [5], which is a part of machine learning [6]. In deep learning workflow, relevant features are automatically extracted from images by a convolutional neural network. Since deep learning performs comprehensive supervised learning [7] where the neural network is given a raw and important data to perform, such as object detection, and learns how to do it automatically, the term deep refers to the number of hidden layers in the neural network.

In general, blind people need to immediately identify the objects to take the appropriate action towards each of them in a timely manner without delay. This requires speed in processing and accuracy in Object definition. Due to that, it is prefer to utilize real-time video stream instead of using a static image. Where, It is captured by the webcam and processed at the same moment, and this provides continuity of streamed data at the receiving side (object detection model).

For real-time streams, it needs more efficient and fast algorithms to extract and derive the specified category. Therefore, some object detection algorithms were developed for object detection such as Region-based Convolution Neural Network (R-CNN, Fast R-CNN, and Faster R-CNN)[9-12], Single Shoot Detector (SSD) [12], and You Only Look Once (YOLO)[13]. There are many versions of YOLO e.g. (YOLO V1, V2, and V3) [14-16], where the efficiency, speed and accuracy differ by the algorithm used.

The difficulties faced by the visually impaired people along with the advances that had been made in object detection algorithms, higher accuracy and efficiency rate that had been achieved. It is inspired us to help the visually impaired people using computer vision to get rid of traditional methods and overcome the dependence of blind people on white canes to move around safely.

In this paper, we have proposed approach to assist visually impaired people by providing audio aid that guides them to percept obstacles in their way. It is using object detection algorithm that quickly and accurately detects objects within images and video streams based on deep neural network called YOLO object detection algorithm. YOLO network is trained and applied to build a software that can detect objects in real-time from the stream captured by the camera and produce a real-time voice response that states the object's identity and location (center, left, right, bottom or top of the camera stream), the objects are recognized such as ( doors, stairs, people, mobile phones) [14]. Spacific dataset (Open images dataset). The following sections in this paper are as follows: Section II related works are discussed. Section III presents a brief description of the proposed approach. Section IV describes the experimental results. Section V is the discussion. In section VI we present the conclusion. In last section VII, we present our future work.

## II. RELATED WORK

Studies which have used deep learning technologies became popular. High-level state-of-art technologies have been developed using computer vision and image processing to help visually impaired people deal with the environment around them, below are the most related ones:

The study in [17] aimed to improving daily life for partially-impaired, blind and visually impaired persons by giving a comprehensive idea about the surrounding indoor environment, they propose indoor object detection system based on deep convolutional neural network (RetinaNet), Resnet as feature extractor and ADAM as network optimizer. It achieves 84.61 mAP at detection precisions.

The purpose of the study in [26] is to propose a smart object detection system in real-time based on convolutional neural network to provide a safe living for visually impaired people. An edge box algorithm was used in SSD algorithm. The trained model was stored in cloud storage database so that the trained model could be retrieved and performing the image recognition process anywhere and anytime and it achieves accuracy of 73.3%, an audio-based detector was generated on the detected object to notify the visually impaired people about the identified object.

A scene perception system for the visually impaired based on object detection and classification using multi-modal DCNN has been introduced in [18]. The system is trained using multicolumn CNN with edges features. The feature maps are classified using CNN having three fully connected layers and one convolutional layer with three fully connected layers respectively, the proposed system is designed for helping visually impaired people. It detects and classifies obstacles that come on the way and the distance of the obstacle from the user and warns the user about that. This model archives 81.5% accuracy in detection.

III. THE PRPOSED APPROACH

This proposed model typically consists of two phases: the first phase is training the YOLO model on a specific dataset (Open images dataset) to obtain weights that can identify and detect objects with good accuracy. The second phase is the implementation of the algorithm using the weights that have been trained to develop a prototype that receives the real-time stream from the camera, detects the objects and produces a voice response with the class name and location of the object as shown in Fig. 1.

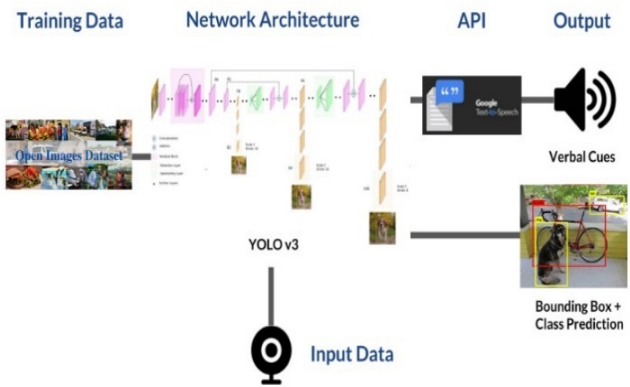

Fig. 1. The proposed detection approach scheme

A. Dataset

In general, the first most important thing in deep learning task is to prepare the dataset. Therefore, a subset of Google's Open Images dataset was used. In addition, OIdv4 toolkit was utilized to download images for four classes of objects that may onstitute an obstacle for the visually impaired, such as doors and stairs or it may be something, they need to find such as a mobile phone or a person. In each category there are nearly 550 image and a total of 2,200 images for the entire used data set. The process of labeling them was manual (Hand-Labeled) using Annotation tool for YOLO. This tool gives the ability to manually draw the bounding boxes for the objects with annotations in a format that YOLO uses during training as follows:

<object-class-id> <center-x> <center-y> <width> <height>

The first field refers to the class id, where the four classes are represented by integers ranging from 0 to number of classes – 1. Here, current case numbers ranges from 0 to 3, the second and third fields refer to the center coordinates of the bounding box, and the fifth and sixth refer to the width and height of the bounding box. Thus, the data set labeled by the YOLO format is ready and after that, we started training this dataset. Afterthought, it is randomly divide the dataset into training data and testing data.

B. YOLO Algorithm Framework

YOLO uses Darknet neural network framework which is fast and supports both CPU and GPU computations. Moreover, it is utilized for object detection and image classification tasks in the form of Convolutional Neural Networks. However, the original repository of this framework only works on Linux and Mac operating systems [22]. Therefore, a forked AlexyAb's repository [23] which is an exact copy of the original Darknet with additional Windows support was used in the training phase.

C. Machine Training Process

1) Transfer Learning :

It is a popular approach in deep learning. Where a pre-trained model developed for a task is reused as the starting point on a second computer vision task instead of learning from scratch. It used to make training process more faster and to improve the performance of the deep learning model. Therefore, a pre-trained model was used which contains only the convolutional layer weights trained on ImageNet dataset (darknet53.conv.74).

2) Preprocessing:

To start model training there are some files that must be prepared first such as object data file, object names file, training images file and testing images file. In addition, YOLOv3 needs a configuration file darknet-YOLOv3.cfg [22]. All the important training parameters are stored in this configuration file.

In this paper, training has been carried out with the next parameters: batch size 64 for batch training, the input data are fed to the network within batches that includes a fixed number of samples. The training images are first resized before training. Here, it uses the values of 608×608 for training. The Learning rate = 0.001 parameter controls how much adjusting the weights of the network. Max_bach = 60000 is the number of iterations required for training and it is determined empirically. At the beginning of the training process, it started with random information and so the learning rate needs to be high. But as the neural network sees a lot of data, the learning rate needs to be decreased over time. In the Cfg file, this decrease is accomplished by first specifying that our learning rate decreasing policy is by steps, and set steps parameter to 80%,90% of max_bach which is 45000,50000. For the data augmentation during network training, some transformations are used such as: changes of hue, saturation, and exposure of the image were in the range 0.1, 1.5, 1.5 respectively. Also filters parameters it is calculated: (classes + 5) x3 = 27 in the convolutional layer before each YOLO layer and change number of classes to 4 class in each YOLO layer.

*3) Training :*

To conduct this step, the training data and the configuration file that contains the parameters and the initial weights are passed. The map factor is used to monitor the accuracy of the model during training. The training process for the model in the iteration 8300 with a mAP in the testing data set. It started from 30 and continued to rise to approximately 80% in the 36 thousand iteration. Where mean Average Precision (mAP) object detection, is a popular metric in measuring the accuracy of object detectors.

*D. System Implementation Process*

The processing of implementation can be summarized as follows:

- The programming languages that support YOLOv3 are limited as it can be implemented using Darknet which is written using C programming language [22]. Since the work is not limited only to detecting objects, it also needs to convert the text into voice. Therefore, the software was executed in Python language the due to its richness in libraries that support computer vision.

- As illustrated in [16], YOLOv3 tends to maintain time efficiency which is the most powerful feature of it.. For the computation of the proposed model, firstly It was imported a library called NumPy which is a basic package for python that provides a powerful N-dimensional array object. In addition, it was imported Open Source Computer Vision (OpenCV) which is a library of python programming functions mainly designed to solve computer vision problems.

- Some parameters are defined such as threshold. That is used to determine the confidence factor in order to generate the bounding box and IOU threshold. After initializing the parameters, labels of the dataset are also defined; paths to trained weights and custom cfg file of my dataset are passed to the loaded YOLO network from darknet in a parameter called Net. As for object detection, YOLO creates boxes and draw each object class with a unique randomly generated color.

- In this model, YOLO creates boxes and draw each object class with a unique randomly generated color. After the input is captured from the camera, a value of the width and the height of the captured data is declared using blob form[24]. After that, the proposed model will be initialized to start making predictions. The result of this process is structured in form of bounding boxes' coordinates, confidence values and labels.

*E. Real Time Voice Feedback Processing*

After capturing real-time video and getting results, text-to-speech conversion is the next task. In this proposed model, the Google's Text to Speech API libraries was used. Which enables to generate speech like human. Furthermore, it converts text to common audio formats such as WAV and MP3 and supports 40 languages.

Firstly, after the package is installed and imported, the audio files are created for the names of all the objects that the model can identify. Then the center bounding box coordinates , height and width of the image and object label, it is passed to a method called play to determine object's location and where the bounding box is located on the screen, is it on the right, left, top, or bottom. It will then give a voice feedback of the location and the name of the detected object to enable the visually impaired person to know the name and the approximate location of the object.

Audio plays for the object detected and its related location using Playsound module which is a python module used for playing sound files, because gTTs API only converts the text to speech and save it as an MP3 file. As known, gTTs API only works online, so all sound files were previously generated and saved offline as MP3 files. Python threading is used to detect objects and give voice feedback at the same time.

## IV. EXPERIMENTAL RESULTS

To evaluate the performance of proposed model, the system was tested under three sets contain different number of images (50, 100, and 230 images). In addition, the number of detected object, the corresponding accuracy ratio and the time taken in those detections were recorded. Since the algorithm resizes the images that processes at a size determined by the user, experiments were performed four times on different sizes of images for each set to obtain the best size that leads to the highest accuracy rate in detecting objects. The results obtained of the proposed approach are showed in Table I.

TABLE I. EXPERIMENTAL RESULTS

| Images count | | 50 | 100 | 230 |
|---|---|---|---|---|
| Size 416*416 | mAP % | 94.23 | 92 | 87.16 |
| | Detect obj | 92 | 187 | 504 |
| | Time sec | 1 | 3 | 8 |
| Size 512*512 | mAP % | 94.50 | 91.90 | 87.32 |
| | Detect obj | 96 | 195 | 533 |
| | Time sec | 2 | 3 | 9 |
| Size 608*608 | mAP % | 97.5 | 92.8 | 90.1 |
| | Detect obj | 106 | 222 | 621 |
| | Time sec | 3 | 5 | 12 |
| Size 832*832 | mAP % | 86.33 | 80.04 | 75.43 |
| | Detect obj | 204 | 415 | 1117 |
| | Time sec | 4 | 9 | 20 |

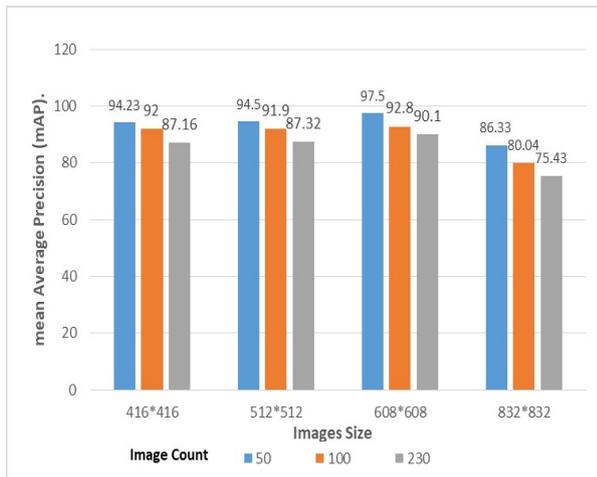

Fig. 2. The accuracy of proposed approach on the datasets.

After experimenting and inspecting the results, it was found that size 608 *608is the best size. As is showed in Fig. 2, it is achieved high accuracy rate when detected 106 object in 3 seconds with 97.50% accuracy mAP for 50 images. Whereas, it detected 222 objects in 5 seconds with 92.87 % accuracy mAP for 100 images, and when using 230 images for testing, the accuracy decreased by only 2.7%, where it achieved 90.17% accuracy mAP in 621 detections in 12 seconds. It was also noted that when using smaller image size, the accuracy was less but the detection speed was faster, and when increasing the size of the images by more than 608 pixels, the accuracy decreased and the speed was slower.

The results obtained in detecting the objects using the model that trained were very encouraging. Its accuracy has been verified in real-time for the purpose that seeking for it in this paper. Fig. 3 illustrates the results of our proposed object detection model in real-time video. Here, we can see the bounded box surrounds a specific object, which means the YOLO model has the capability to identify object from its surrounding. In addition, it has the capability to recognize objects even it did not see before as long as they have the same features of objects which previously trained on during the learning process. Where, it will find the most appropriate confidence factor, and then it represents the label as shown in the Fig. 3.

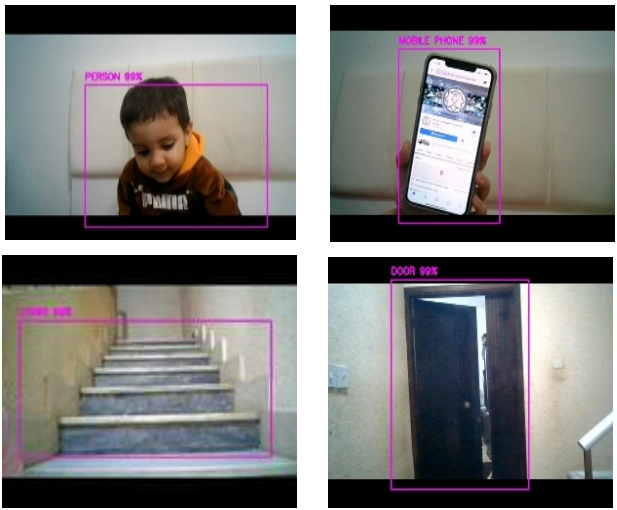

Fig. 3. Detected object with label and accuracy rate.

As addresses in Fig. 3, the object's label is shown correctly and its location is surrounded with a bounding box that means the identification has been done. The accuracy rate of the object identification is also showed next to the label of. As we mention before, gTTs Google's ApI was used for text to audio processing to so that the visually impaired people can hear audio output of class names of objects that were recognized through real-time object detection process. Unfortunately, cannot show this text -to-audio process in this paper as it can only be shown in practical implementation which it has done and witnessed that it was successful.

## V. DISSCUTION

Previous studies [17], [18], [26] similar to the proposed approach aimed to help blind people whereas they have used other algorithms and techniques. Performance comparison of YOLO real-time detection model according to accuracy and real time with other these techniques is addressed in Table II.. Our model results were similar to them. This indicates the efficiency and speed of the performance of our YOLOv3 model to detected objects in real time.

TABLE II. THE PERFORMANCE COMPARISON WITHE OTHER ALGORITHMS IN PREVIOUS STUDIES

| Algorithm | Accuracy | Real time |
|---|---|---|
| RetinNet+Resnet | 84.6% | NO |
| SSD+edge box | 73.3% | Yes |
| DCNN | 81.5% | NO |
| YOLOv3 | 90.17% | Yes |

In addition, the results of the proposed model trained on the custom dataset were compared with the YOLOv3 models previously trained on other datasets according to their research [22], [25]. As, the results obtained through training proposed model are considered excellent as addresses in Table. III. As a high percentage of recognition was obtained in detecting the objects are up to 90%. On other hand, the number of objects that proposed model can identify, it has been trained on four classes only to reduce the training time and utilize the time in the optimization process. However, what has been applied on four classes can be easily applied on any possible number of objects with the availability of the data set and the resources needed for the training process such as a powerful graphics card etc.

TABLE III. THE PERFORMANCE COMPARISON WITH OTHER DATASET IN PREVIOUS STUDIES

| Dataset | Input | Classes | mAP% |
|---|---|---|---|
| COCO [22] | 608 | 80 | 57.9% |
| KITTI [25] | 512 | 3 | 82.95% |
| Pascal voc2012[25] | 416 | 20 | 79.26% |
| Proposed approach | 608 | 4 | 90.17% |
| | 512 | | 87.32% |
| | 416 | | 87.16% |

## VI. CONCLUSION

This paper presents detection and identification of objects model to assist visually impaired people. For the object recognition process, the YOLOv3 algorithm was trained where we have implemented deep learning process using Darknet framework as well as OpenCV libraries of Python. In addition, Web camera was used to capture live video stream. After that, the objects would be recognized via YOLOv3 model and then objects detection results would be transferred to the visually impaired user by means of text to speech API using gTTs library. Experimental results demonstrate that the proposed model achieved an accuracy rate of 90% in testing dataset and it could sometimes reach 100% in real-time detection.

## VII. Future Directions

For future studies, there are several things we want to try to make the application more useful for the visually impaired user. First, we plane to determine the depth of the detected object to know the distance the object is away from the visually impaired user. Further, we will study how to design the hardware architecture to make it a portable, easy-to-use device with interfaces designed for voice commands for visually impaired people.

REFERENCES

[1] World Health Organisation, "Global Data on," Glob. Data Vis. Impair. 2010, p. 17, 2010, [Online]. Available: http://www.who.int/blindness/GLOBALDATAFINALforweb.pdf.

[2] Jan Solem, Programming Computer Vision with Python: Tools and Algorithms for Analyzing Images. 2012.

[3] J. D. Prince, Computer Vision: Models, Learning, and Inference. 2012.

[4] V. Wiley and T. Lucas, "Computer Vision and Image Processing: A Paper Review," Int. J. Artif. Intell. Res., vol. 2, no. 1, p. 22, 2018, doi: 10.29099/ijair.v2i1.42.

[5] A. Mathew, P. Amudha, and S. Sivakumari, "Deep learning techniques: an overview," Adv. Intell. Syst. Comput., vol. 1141, no. August 2020, pp. 599–608, 2021, doi: 10.1007/978-981-15-3383-9_54.

[6] H. Wang, C. Ma, and L. Zhou, "A brief review of machine learning and its application," Proc. - 2009 Int. Conf. Inf. Eng. Comput. Sci. ICIECS 2009, 2009, doi: 10.1109/ICIECS.2009.5362936.

[7] M. Bojarski et al., "End to End Learning for Self-Driving Cars," pp. 1–9, 2016, [Online]. Available: http://arxiv.org/abs/1604.07316.

[8] L.G Shapiro, G.C. Stockman, "Imaging and Image Representation," Computer Vision (March 2000), http://www.cse.msu.edu/~stockman/Book/book.html.

[9] R. Girshick, J. Donahue, T. Darrell, and J. Malik, "Rich feature hierarchies for accurate object detection and semantic segmentation," Proc. IEEE Comput. Soc. Conf. Comput. Vis. Pattern Recognit., pp. 580–587, 2014, doi: 10.1109/CVPR.2014.81.

[10] J. R. R. Uijlings, K. E. A. Van De Sande, T. Gevers, and A. W. M. Smeulders, "Selective search for object recognition," Int. J. Comput. Vis., vol. 104, no. 2, pp. 154–171, 2013, doi: 10.1007/s11263-013-0620-5.

[11] R. Girshick, "Fast R-CNN," Proc. IEEE Int. Conf. Comput. Vis., vol. 2015 Inter, pp. 1440–1448, 2015, doi: 10.1109/ICCV.2015.169.

[12] S. Ren, K. He, R. Girshick, and J. Sun, "Faster R-CNN: Towards Real-Time Object Detection with Region Proposal Networks," IEEE Trans. Pattern Anal. Mach. Intell., vol. 39, no. 6, pp. 1137–1149, 2017, doi: 10.1109/TPAMI.2016.2577031.

[13] W. Liu et al., "SSD: Single shot multibox detector," Lect. Notes Comput. Sci. (including Subser. Lect. Notes Artif. Intell. Lect. Notes Bioinformatics), vol. 9905 LNCS, pp. 21–37, 2016, doi: 10.1007/978-3-319-46448-0_2.

[14] J. Redmon, S. Divvala, R. Girshick, and A. Farhadi, "You only look once: Unified, real-time object detection," Proc. IEEE Comput. Soc. Conf. Comput. Vis. Pattern Recognit., vol. 2016-Decem, pp. 779–788, 2016, doi: 10.1109/CVPR.2016.91.

[15] J. Redmon and A. Farhadi, "YOLO9000: Better, faster, stronger," Proc. - 30th IEEE Conf. Comput. Vis. Pattern Recognition, CVPR 2017, vol. 2017-Janua, pp. 6517–6525, 2017, doi: 10.1109/CVPR.2017.690.

[16] J. Redmon and A. Farhadi, "YOLOv3: An incremental improvement," arXiv, 2018.

[17] M. Afif, R. Ayachi, Y. Said, E. Pissaloux, and M. Atri, "An Evaluation of RetinaNet on Indoor Object Detection for Blind and Visually Impaired Persons Assistance Navigation," Neural Process. Lett., vol. 51, no. 3, pp. 2265–2279, 2020, doi: 10.1007/s11063-020-10197-9.

[18] B. Kaur and J. Bhattacharya, "Scene perception system for visually impaired based on object detection and classification using multimodal deep convolutional neural network", J. Electron. Imaging, vol. 28, no. 01, p. 1, 2019, doi: 10.1117/1.jei.28.1.013031.

[19] F. Sultana, A. Sufian, and P. Dutta, "A review of object detection models based on convolutional neural network," arXiv, pp. 11104–11109, 2019.

[20] S. Albawi, T. A. M. Mohammed, and S. Alzawi, "Layers of a Convolutional Neural Network," Ieee, 2017, [Online]. Available: https://wiki.tum.de/display/lfdv/Layers+of+a+Convolutional+Neural+Network.

[21] V. Sangeetha and K. J. R. Prasad, "Syntheses of novel derivatives of 2-acetylfuro[2,3-a]carbazoles, benzo[1,2-b]-1,4-thiazepino[2,3-a]carbazoles and 1-acetyloxycarbazole-2- carbaldehydes," Indian J. Chem. - Sect. B Org. Med. Chem., vol. 45, no. 8, pp. 1951–1954, 2006, doi: 10.1002/chin.200650130.

[22] Joseph Redmon, "Darknet Framework." https://pjreddie.com/darknet/.

[23] Alexyab, "forked Darknet." https://github.com/AlexeyAB/darknet.

[24] B. Kang, "A Review on Image & Video Processing," no. August, 2014.

[25] J. Moran, L. Haibo, Z. Wang, H. Bin, and C. Zheng, ''The application of improved YOLO V3 in multi-scale target detection,'' Appl. Sci., vol. 9, no. 18, pp. 3775–3788, Sep. 2019.

[26] Y. Wong, J. Lai, S. Ranjit, A. Syafeeza, and N. Hamid, (2019). Convolutional neural network for object detection system for blind people. Journal of Telecommunication, Electronic and Computer Engineering (JTEC), 11(2), vol. 11, no. 2, p. 6, 2019.